\documentclass[10pt,twocolumn,letterpaper]{article}

\usepackage{cvpr}              

\newcommand{\red}[1]{{\color{red}#1}}

\newcommand{\blue}[1]{{\color{blue}#1}}

\definecolor{LightGray}{gray}{0.9}
\definecolor{LightGreen}{rgb}{0.9, 1.0, 0.9}

\usepackage{amsmath,amssymb,amsfonts}
\usepackage{times}
\usepackage{epsfig}
\usepackage{algorithmic}
\usepackage{graphicx}
\usepackage{textcomp}
\usepackage{booktabs}
\usepackage{colortbl}
\usepackage{xcolor}
\usepackage{siunitx}
\usepackage{makecell}
\usepackage{bm}
\usepackage{multirow}
\usepackage{adjustbox}
\usepackage{lipsum}
\usepackage{float}
\usepackage{afterpage}
\usepackage{comment}
\usepackage{cuted}
\usepackage{capt-of}
\usepackage{caption}
\usepackage{stfloats}
\usepackage{kotex}
\usepackage{adjustbox}

\definecolor{GoodColor}{HTML}{228B22} 
\definecolor{BadColor}{HTML}{D70000}  
\definecolor{PurpleBox}{HTML}{7030A0}  

\newcommand{\good}[1]{\scriptsize\textcolor{GoodColor}{(#1)}}
\newcommand{\bad}[1]{\scriptsize\textcolor{BadColor}{(#1)}}

\newcommand{\ul}[1]{\underline{#1}}

\definecolor{cvprblue}{rgb}{0.21,0.49,0.74}
\usepackage[pagebackref,breaklinks,colorlinks,allcolors=cvprblue]{hyperref}

\def\paperID{24} 
\def\confName{CVPR}
\def\confYear{2026}

\title{VDPP: Video Depth Post-Processing for Speed and Scalability}

\author{
    Daewon Yoon$^{1,2,*}$ \quad
    Injun Baek$^{1,2,*}$ \quad
    Sangyu Han$^{1}$ \quad
    Yearim Kim$^{1}$ \quad
    Nojun Kwak$^{1,\dagger}$\\[0.6em]
    \textsuperscript{1} Seoul National University \qquad
    \textsuperscript{2} Samsung Electronics\\[0.2em]
    $^{*}$ Equal contribution \qquad
    $^{\dagger}$ Corresponding author
}

\begin{document}
\maketitle
\begin{strip}
  \centering
  \includegraphics[width=.9\linewidth]{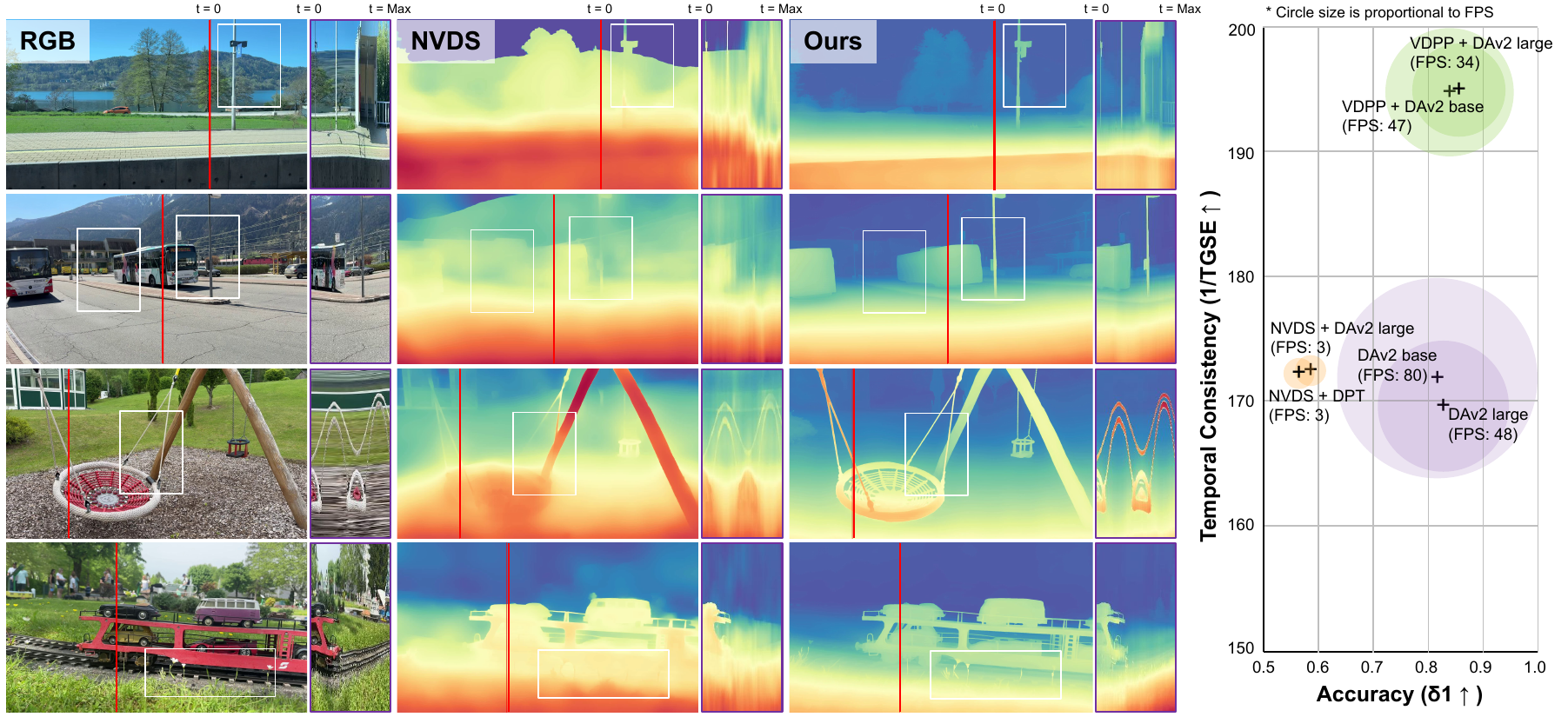}
  \captionsetup{type=figure, width=1.00\linewidth}
  \caption{\textbf{VDPP} achieves superior speed, spatial accuracy and temporal consistency among post-processing methods. \textbf{Left:} Qualitative comparison on four challenging dynamic scenes (top to bottom: scenery viewed from a train, passing buses at a station, a swing with depth variation, and a car carrier trailer). The temporal profile \textcolor{PurpleBox}{purple box} is a slit-scan image by sequentially stacking the \red{red vertical pixel line} over time. While NVDS~\cite{nvds} suffers from severe flickering artifacts and degraded spatial fidelity, our \textbf{VDPP} preserves intricate geometric details while maintaining a stable temporal profile. \textbf{Right:} Quantitative comparison evaluating the temporal consistency, accuracy, and speed on the Sintel dataset~\cite{sintel} at a resolution of 384$\times$384. This comparison uses DAv2~\cite{depthanythingv2} as the image-depth backbone for a fair comparison. VDPP significantly enhances both spatial accuracy and temporal stability while achieving fast inference speed ($>43.5$ FPS), outperforming existing post-processing frameworks by an order of magnitude. Circle sizes indicate inference speed(FPS).}
  \label{fig:teaser} 
\end{strip}
 
\begin{abstract}
Video depth estimation is essential for providing 3D scene structure in applications ranging from autonomous driving to mixed reality. Current end-to-end video depth models have established state-of-the-art performance. Although current end-to-end (E2E) models have achieved state-of-the-art performance, they function as tightly coupled systems that suffer from a significant adaptation lag whenever superior single-image depth estimators are released.
To mitigate this issue, post-processing methods such as NVDS offer a modular plug-and-play alternative to incorporate any evolving image depth model without retraining. However, existing post-processing methods still struggle to match the efficiency and practicality of E2E systems due to limited speed, accuracy, and RGB reliance.

In this work, we revitalize the role of post-processing by proposing \textbf{VDPP (Video Depth Post-Processing)}, a framework that improves the speed and accuracy of post-processing methods for video depth estimation. By shifting the paradigm from computationally expensive scene reconstruction to targeted geometric refinement, VDPP operates purely on geometric refinements in low-resolution space. This design achieves exceptional speed ($>43.5$ FPS on NVIDIA Jetson Orin Nano) while matching the temporal coherence of E2E systems, with dense residual learning driving geometric representations rather than full reconstructions. Furthermore, our VDPP’s RGB-free architecture ensures true scalability, enabling immediate integration with any evolving image depth model. Our results demonstrate that VDPP provides a superior balance of speed, accuracy, and memory efficiency, making it the most practical solution for real-time edge deployment. Our project page is at \href{https://github.com/injun-baek/VDPP}{https://github.com/injun-baek/VDPP}

\end{abstract}
\section{Introduction}
Video depth estimation, the task of inferring 3D scene structure from 2D video sequences, is important 
for a variety of modern applications such as autonomous driving, robotics, and mixed reality. Driven by this critical need, monolithic end-to-end (E2E) models, such as Video Depth Anything (VDA)~\cite{vda}, have established the current state-of-the-art, delivering exceptional coherence and spatial accuracy. 

However, as summarized in Table~\ref{tab:comparison}, current video depth paradigms involve distinct trade-offs for practical deployment. High-performance E2E models achieve benchmarks for spatial accuracy and temporal stability but are typically too computationally intensive for real-time use on edge devices. Furthermore, their tightly coupled design lead to a significant adaptation lag; whenever a superior single I2D estimator is released, it cannot be easily integrated into existing E2E frameworks without full-scale, computationally expensive retraining. A straightforward alternative of applying I2D estimators frame-by-frame offers high spatial accuracy, but results in severe flickering artifacts due to a lack of inherent temporal coherence.

To mitigate these issues, modular post-processing methods like NVDS~\cite{nvds} have emerged as a promising plug-and-play alternative. By decoupling the video stabilization process from the image-level depth estimation, these methods allow for the immediate integration of evolving I2D models. Despite their potential, they have yet to match the efficiency of E2E systems. As illustrated in Figure~\ref{fig:teaser}, NVDS still suffers from severe flickering and degraded spatial fidelity, requiring orders of magnitude longer inference time and relying on RGB data.

In this work, we revitalize the role of post-processing in video depth estimation by reframing video stabilization as a targeted geometric correction task.
We observe that the per-frame depth maps already provide a fundamentally sound, albeit temporally unstable, 3D structure.
Thus, we reframe the task from a video reconstruction problem into a more efficient structure correction problem. Instead of rebuilding the 3D scene from color context, our approach focuses on rectifying unstable elements using geometric cues.

As shown in the quantitative evaluation in Figure~\ref{fig:teaser}~(Right), VDPP significantly enhances both spatial accuracy and temporal stability while achieving the fastest inference speed ($>43.5$ FPS), outperforming existing post-processing frameworks by an order of magnitude.
\begin{table}[t]
\centering
\caption{\textbf{Comparison of video depth estimation paradigms.} VDPP provides a balanced profile for practical deployment, achieving SOTA-level coherence within a modular and efficient post-processing framework.}
\label{tab:comparison}
\small
\setlength{\tabcolsep}{5pt}
\resizebox{\linewidth}{!}{
\begin{tabular}{l|l|ccccc}
\toprule
\textbf{Category} & \textbf{Method} & \textbf{Speed} & \textbf{Spatial} & \textbf{Temporal} & \textbf{Plug \&} & \textbf{Light-} \\
 & & \textbf{($>30$ FPS)} & \textbf{Accuracy} & \textbf{Coherence} & \textbf{Play} & \textbf{weight} \\
\midrule
\multirow{2}{*}{End-to-End} & VDA-S~\cite{vda} & \checkmark & \checkmark & $\triangle$ & $\times$ & $\times$ \\
 & VDA-L~\cite{vda} & $\times$ & \checkmark & \checkmark & $\times$ & $\times$ \\
\midrule
Image-to-Depth & DAv2~\cite{depthanythingv2} & \checkmark & \checkmark & $\times$ & - & \checkmark \\
\midrule
\multirow{2}{*}{Post-Processing} & NVDS~\cite{nvds} & $\times$ & $\times$ & $\triangle$ & \checkmark & $\times$ \\
 & \textbf{VDPP (Ours)} & \checkmark & \checkmark & \checkmark & \checkmark & \checkmark \\
\bottomrule
\end{tabular}}
\raggedright
\footnotesize
* Lightweight capability is verified on the NVIDIA Jetson Orin Nano (8GB) without Out-of-Memory errors.
\vspace{-4mm}
\end{table}
Our primary contributions are summarized as follows:
\begin{itemize}
\item \textbf{Decoupled and Scalable Paradigm:} We propose an RGB-free, modular framework that resolves the adaptation lag of monolithic models, enabling the immediate integration of any evolving image-depth backbone or depth-only sensors such as LiDAR.

\item \textbf{Embedded Efficiency and Real-time Deployment:} Designed for resource-constrained environments, VDPP achieves exceptional speed ($>43.5$ FPS) on an NVIDIA Jetson Orin Nano with a minimal memory footprint, ensuring viability for real-time edge applications. 

\item \textbf{Geometric Refinement with Residual Learning:} We introduce a dual-component design (Depth + Normals) with downsampling and residual learning in error space. This technical shift focuses the network on targeted geometric correction rather than redundant scene reconstruction, allowing a lightweight architecture to match the temporal coherence of heavy end-to-end systems.
\end{itemize}
\vspace{3mm}
\section{Related Work}
Our work adopts a post-processing perspective to video depth estimation, aiming to complement the spatial accuracy of modern image-to-depth (I2D) models with a lightweight mechanism for temporal stability. By utilizing a depth-only refinement strategy, we seek to address the inherent trade-offs between computational efficiency and temporal coherence.

\subsection{Image-to-Depth Estimation}
Early deep learning-based approaches~\cite{rel_early_depth_2014, rel_early_depth2_2018, rel_early_depth3_2019} to image-to-depth~(I2D) estimation showed remarkable progress, driven by architectures adept at understanding monocular cues.
The introduction of Vision Transformers marked a significant turning point, as these architectures~\cite{rel_single_image1_2021, rel_single_image4_2023depthformer, rel_single_image5_2022, rel_single_images6_2021, rel_single_image7_2020midas} began to demonstrate high effectiveness for depth estimation.
Architectures like DPT~\cite{dpt} pioneered their use for dense prediction tasks by effectively capturing global scene context.
Further advances led to models such as Depth Anything V2~\cite{depthanything, depthanythingv2}, which leverages large-scale data and efficient architectures to achieve exceptional performance across various scenarios.
While these models offer impressive speed and spatial accuracy for single images, their naive frame-by-frame application to video results in severe flicker artifacts due to a lack of temporal coherence. 
This limitation motivates our proposed \textit{VDPP}, which preserves the efficiency and fidelity of I2D models while enforcing temporal coherence via depth-only post-processing.

\subsection{Video Depth Estimation}
Video depth estimation is to generate temporally consistent depth maps by eliminating visual discrepancies between consecutive frames, `flickering'.
Test-time training strategies~\cite{rel_testtime1, rel_testtime2, rel_testtime3, rel_testtime4} were predominant in early attempts to enhance temporal consistency.
This strategy involves adapting a pre-existing I2D model to an input video during the inference stage.
However, it demands significant computational resources and often struggles to guarantee stable performance in unpredictable scenarios.
Subsequently, end-to-end architectures emerged that process video clips directly. These models adopt various architectures, such as transformer adapters (ViTA~\cite{vita2023}), spatio-temporal attention (VDA~\cite{vda}), or feature warping based on optical flow (MAMo~\cite{mamo2023}). Recently, diffusion-based methods, such as ChronoDepth~\cite{chronodepth2025}, DepthCrafter~\cite{depthcrafter2025}, and DepthAnyVideo~\cite{depthanyvideo2024}, have gained attention for their performance in detail and consistency.
Although these models produce excellent results, they face remarkably slow inference speeds and substantial memory requirements.

\subsection{Post-Processing Approaches.}
Post-processing methods such as NVDS and NVDS+~\cite{nvds, nvdsplus} introduced a plug-and-play paradigm by stabilizing pre-computed depth maps through bidirectional inference and large-scale training. Despite their convenience, they exhibit two key weaknesses. First, their reliance on RGB inputs undermines applicability in privacy-sensitive or non-RGB settings (e.g., LiDAR, ToF). Second, their refinement is limited to coarse global corrections, offering insufficient local accuracy while adding computational overhead.
\vspace{3mm}
\section{Method}
\begin{figure*}[ht!]
    \captionsetup{justification=justified, singlelinecheck=false}
    \centering
    \includegraphics[width=\linewidth]{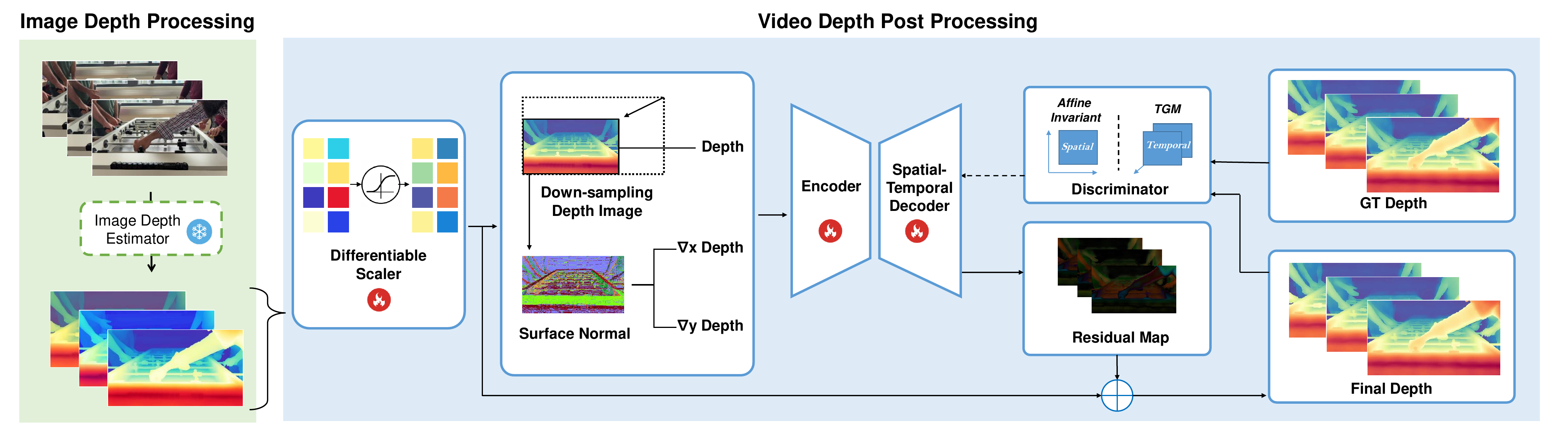}     
    \caption{\textbf{The architectural overview of VDPP.} Our post-processing pipeline takes per-frame depth maps from \textbf{any Image Depth Estimator} as input. The pipeline begins with using \textbf{a learnable Differentiable Scaler} based on the median to normalize the scale of input depth maps. The normalized map is used to create a \textbf{3-channel geometric manifold through down-sampling}. A transformer encoder (DINOv2~\cite{dinov2}) and Spatial-Temporal Decoder (VDA~\cite{vda}) predict \textbf{a residual map}, which is then added back to the scaled input via a shortcut connection to produce the final depth. Training is supervised by comparing final depth with the ground truth depth using a composite temporal (TGM) and spatial (Affine Invariant) loss function~\cite{vda}.}
    \label{fig:architecture}
\end{figure*}
Our proposed framework, VDPP, introduces a modular paradigm for video depth estimation by transitioning from global scene reconstruction to intelligent spatio-temporal refinement.
Instead of generating depth from scratch, VDPP utilizes existing per-frame depth maps—which often possess high spatial fidelity but low temporal stability—and refines them into a coherent and accurate video sequence. 
This refinement process leverages three core principles: adaptive standardization, efficient geometric representation learning, and precise error modeling.
The overall pipeline is depicted in Figure~\ref{fig:architecture}.

\subsection{Adaptive Stabilization Framework}
\label{subsec:adaptive_framework}
The initial stage addresses a primary source of temporal instability: global scale variations. 
To address the scale variations inherent in per-frame depth estimation, we introduce a learnable normalization module, the \textbf{Differentiable Scaler} ($\mathcal{H}_{\text{scale}}$).
For each initial depth map $D_{\text{initial}}^{(t)}$ ($t$ denotes the frame index.), this scaler uses the median $m$ over the mean for its robustness to outliers.
For example, a single sky pixel with near infinite depth cannot skew the median as it would the mean.
This representative statistic is then passed to our learnable module, $\mathcal{H}_{\text{scale}}$, which predicts an adaptive depth scale factor:
\begin{equation}
    D_{\text{scaled}}^{(t)} = \mathcal{H}_{\text{scale}}\left( \underset{i,j}{\text{median}}(D_{\text{initial}}^{(t)}[i,j]) \right) \cdot D_{\text{initial}}^{(t)}
\end{equation}
where $\mathcal{H}_{\text{scale}}(m) = \exp(\tanh(- a \cdot m + b))$, $(i,j)$ are the pixel coordinates, and $a, b$ are learnable parameters.
This process preemptively handles temporal flickering by softly adjusting the global scale of each frame.


\subsection{Efficient Downsampling Framework}
\label{subsec:geometric_downsampling}
Unlike RGB-based models that require high-resolution inputs to disentangle geometry from appearance (e.g., lighting and texture), we enable efficient processing through two key insights.
Although aggressive downsampling inevitably loses fine spatial details, we mitigate this via fine geometric representaion and residual learning.
we construct a fine geometric manifold by concatenating depth (1D) with surface normals (2D), which are the spatial gradients $\nabla_x$ and  $\nabla_y$ of depth representation.

\subsubsection{Downsampling strategy}
Since our input is an existing per-frame depth map $D_{\text{scaled}}^{(t)}$ , we can utilize aggressive downsampling via bilinear interpolation at a ratio (0.5), with minimal loss of structural information. By performing feature extraction and temporal attention at a lower spatial resolution, we significantly reduce the computational burden on the transformer encoder:
\begin{equation}
    D_{\text{down}}^{(t)} = \operatorname{Downsample}_{r}(D_{\text{scaled}}^{(t)})
\end{equation}

\subsubsection{Geometric Representation}
To preserve critical structural details during downsampling, we construct a 3-channel geometric manifold by concatenating depth with its spatial gradients ($\nabla_x, \nabla_y$) extracted using a Sobel filter, which represent surface normals:
\begin{equation}
    I_{\text{down}}^{(t)} = \operatorname{concat}(D_{\text{down}}^{(t)}, \nabla_x D_{\text{down}}^{(t)}, \nabla_y D_{\text{down}}^{(t)})
\end{equation}
This representation allows the network to focus on local surface orientation, which facilitates faster convergence and improved accuracy without the need to process appearance-based cues such as lighting and texture.


\subsection{Refinement Network and Residual Prediction}
\label{subsec:refinement}
The final stage performs the detailed correction, using \textbf{Learnable Refinement Network} and \textbf{Residual Prediction}. Specifically, features extracted by a DINOv2~\cite{dinov2} backbone, serving as the encoder, are passed to a spatial-temporal decoder adapted from VDA~\cite{vda}. We denote this decoder function, which utilizes its inherent temporal attention mechanism, as $\mathcal{H}_{\text{temporal}}$. Crucially, the encoder is trained to extract a feature representation $\mathbf{f}_{t}$ from each input $I_{\text{down}}^{(t)}$. The decoder $\mathcal{H}_{\text{temporal}}$ then fuses a temporal window of these features. $k$ is the window size (e.g., $k=16$) to directly output the desired per-pixel \textbf{residual correction}, $R^{(t)}$:
\begin{equation}
    R^{(t)} = \mathcal{H}_{\text{temporal}}(\mathbf{f}_{t}, \dots, \mathbf{f}_{t+k})
\end{equation}
This learned residual map is added to the scaled depth map from the first stage, $D_{\text{scaled}}^{(t)}$, to produce the final output, $\hat{D}^{(t)}$.
\begin{equation}
    \hat{D}^{(t)} = D_{\text{scaled}}^{(t)} + R^{(t)}
\end{equation}
This paradigm simplifies the learning task. By focusing on residual corrections rather than full reconstruction, the network benefits from a smoother optimization landscape and can concentrate its capacity on correcting subtle artifacts.

\subsection{Training Objective}
The network is optimized via a unified spatio-temporal loss:
\textbf{Spatial Loss} $\mathcal{L}_{\text{spatial}}$:An affine-invariant loss that unifies scale-shift matching, following the approach proposed in MiDaS~~\cite{rel_single_image7_2020midas}".
\begin{equation}
\mathcal{L}_{\text{spatial}} = \frac{1}{HW}\sum_{i=1}^{H}\sum_{j=1}^{W}\rho\big(\hat D(i,j),\, D(i,j)\big)
\end{equation}
where $H$ and $W$ denote the image height and width, and $\rho$ is the affine-invariant absolute error; $D^{(t)}(i,j)$ and $\hat{D}^{(t)}(i,j)$ denote the ground-truth and predicted depths at pixel $(i, j)$.
\textbf{Temporal Loss} $\mathcal{L}_{\text{temporal}}$: We adopt the Temporal Gradient Matching (TGM) loss introduced by VDA~\cite{vda}. TGM compares frame-to-frame changes between predicted and ground-truth sequences across multiple temporal scales within the sequence. By minimizing the difference between these temporal gradients over both short and long intervals, the model learns to produce depth videos where motion and temporal variations appear consistent with the ground truth.
\begin{equation}
\mathcal{L}_{\text{temporal}} = \frac{1}{T-1}\sum_{t=1}^{T-1}\left\|\nabla_t \hat{D} - \nabla_t D\right\|_1
\end{equation}
where $\nabla_t$ denotes the temporal gradient operator and $T$ is the sequence length. The final training objective combines both spatial and temporal components:
\begin{equation}
    \mathcal{L}_{\text{total}} = \alpha \mathcal{L}_{\text{spatial}} + \beta \mathcal{L}_{\text{temporal}}
\end{equation}
where $\alpha$ and $\beta$ are weights that control the trade-off between spatial accuracy and temporal consistency.

\section{Experiments}

\begin{table*}[hb!]
\centering
\caption{\textbf{Inference speed comparison}. Existing post-processing is slow and performs poorly, our VDPP shows superior speed. Notably, \textbf{DAv2-B + VDPP and DPT-L + VDPP achieve superior speed among video depth estimation methods.} Speed is measured in Frames Per Second (FPS) and time components in milliseconds (ms). Performance metrics are reported in Table \ref{tab:performance}. \textbf{Bold numbers} indicate the best performance, and \ul{underlined numbers} represent the second-best, excluding single-frame image to depth methods.}
\label{tab:speed}
\setlength{\tabcolsep}{4pt} 
\resizebox{\textwidth}{!}{ 
\begin{tabular}{@{}lrrrrrrrrrrrr@{}}
\toprule
\multirow{2}{*}{\textbf{Method}} & \multicolumn{4}{c|}{NYUv2~\cite{nyuv2}} & \multicolumn{4}{c|}{Bonn~\cite{bonn}} & \multicolumn{4}{c}{Sintel~\cite{sintel}} \\ \cmidrule(l){2-13} 
 & \multicolumn{1}{c}{FPS ↑} & \multicolumn{1}{c}{\begin{tabular}[c]{@{}c@{}}Total\\ (ms)\end{tabular}} & \multicolumn{1}{c}{\begin{tabular}[c]{@{}c@{}}Image to\\ Depth (ms)\end{tabular}} & \multicolumn{1}{c}{\begin{tabular}[c]{@{}c@{}}Depth to\\ Video (ms)\end{tabular}} & \multicolumn{1}{c}{FPS ↑} & \multicolumn{1}{c}{\begin{tabular}[c]{@{}c@{}}Total\\ (ms)\end{tabular}} & \multicolumn{1}{c}{\begin{tabular}[c]{@{}c@{}}Image to\\ Depth (ms)\end{tabular}} & \multicolumn{1}{c}{\begin{tabular}[c]{@{}c@{}}Depth to\\ Video (ms)\end{tabular}} & \multicolumn{1}{c}{FPS ↑} & \multicolumn{1}{c}{\begin{tabular}[c]{@{}c@{}}Total\\ (ms)\end{tabular}} & \multicolumn{1}{c}{\begin{tabular}[c]{@{}c@{}}Image to\\ Depth (ms)\end{tabular}} & \multicolumn{1}{c}{\begin{tabular}[c]{@{}c@{}}Depth to\\ Video (ms)\end{tabular}} \\ \midrule
\textit{End-to-end} & \multicolumn{1}{l}{} & \multicolumn{1}{l}{} & \multicolumn{1}{l}{} & \multicolumn{1}{l}{} & \multicolumn{1}{l}{} & \multicolumn{1}{l}{} & \multicolumn{1}{l}{} & \multicolumn{1}{l}{} & \multicolumn{1}{l}{} & \multicolumn{1}{l}{} & \multicolumn{1}{l}{} & \multicolumn{1}{l}{} \\
VDA-S & 14 & 72.7 & {---} & {---} & 60 & 16.7 & {---} & {---} & 53 & 18.8 & {---} & {---} \\
VDA-L & 1 & 669.1 & {---} & {---} & 9 & 109.3 & {---} & {---} & 6 & 166.2 & {---} & {---} \\ \midrule
\textit{Image to Depth} & \multicolumn{1}{l}{} & \multicolumn{1}{l}{} & \multicolumn{1}{l}{} & \multicolumn{1}{l}{} & \multicolumn{1}{l}{} & \multicolumn{1}{l}{} & \multicolumn{1}{l}{} & \multicolumn{1}{l}{} & \multicolumn{1}{l}{} & \multicolumn{1}{l}{} & \multicolumn{1}{l}{} & \multicolumn{1}{l}{} \\
DPT-L & 82 & 12.2 & 12.2 & {---} & 85 & 11.8 & 11.8 & {---} & 63 & 15.9 & 15.9 & {---} \\
DAv2-B & 80 & 12.5 & 12.5 & {---} & 75 & 13.3 & 13.3 & {---} & 72 & 13.8 & 13.8 & {---} \\
DAv2-L & 48 & 20.8 & 20.8 & {---} & 47 & 21.5 & 21.5 & {---} & 38 & 26.1 & 26.1 & {---} \\ \midrule
\textit{Post-processing} & \multicolumn{1}{l}{} & \multicolumn{1}{l}{} & \multicolumn{1}{l}{} & \multicolumn{1}{l}{} & \multicolumn{1}{l}{} & \multicolumn{1}{l}{} & \multicolumn{1}{l}{} & \multicolumn{1}{l}{} & \multicolumn{1}{l}{} & \multicolumn{1}{l}{} & \multicolumn{1}{l}{} & \multicolumn{1}{l}{} \\
DPT-L + NVDS & 3 & 367.5 & 12.2 & 355.3 & 5 & 218.6 & 11.8 & 206.8 & 2 & 554.8 & 15.9 & 538.8 \\
DAv2-B + NVDS & 3 & 367.8 & 12.5 & 355.3 & 5 & 220.2 & 13.3 & 206.8 & 2 & 552.7 & 13.8 & 538.8 \\
DAv2-L + NVDS & 3 & 376.1 & 20.8 & 355.3 & 4 & 228.3 & 21.5 & 206.8 & 2 & 564.9 & 26.1 & 538.8 \\
\textbf{DPT-L + Ours} & \textbf{48} & \textbf{20.8} & 12.2 & 8.6 & \textbf{76} & \textbf{13.2} & 11.8 & 1.4 & {\ul {56}} & {\ul {18.0}} & 15.9 & 2.0 \\
\textbf{DAv2-B + Ours} & {\ul{47}} & {\ul {21.1}} & 12.5 & 8.6 & {\ul {68}} & {\ul {14.7}} & 13.3 & 1.4 & \textbf{63} & \textbf{15.9} & 13.8 & 2.0 \\
\textbf{DAv2-L + Ours} & 34 & 29.4 & 20.8 & 8.6 & 44 & 22.9 & 21.5 & 1.4 & 36 & 28.1 & 26.1 & 2.0 \\ \bottomrule
\end{tabular}
} 
\end{table*}

We conducted comprehensive experiments to validate VDPP's ability to simultaneously achieve speed, accuracy, and scalability. Our evaluation spans three diverse benchmarks that cover static, dynamic, and synthetic datasets, with testing on both high-performance and edge hardware. We demonstrate that VDPP not only matches state-of-the-art end-to-end models in spatial accuracy and temporal consistency while achieving superior speed, but also uniquely enables deployment on memory-constrained devices and direct processing of depth sensors like LiDAR. Finally, we conduct several ablation studies with ablated models.

\subsection{Datasets and Evaluation Metrics}
For quantitative evaluation, we use a diverse set of three standard benchmarks to cover the static (NYUv2~\cite{nyuv2}), dynamic (Bonn~\cite{bonn}), and synthetic (Sintel~\cite{sintel}) scenarios. All experiments are conducted by sampling 16 frames per video, and evaluating on both the original image resolutions and a 384x384 small resolution for real-time performance. For qualitative evaluation, we additionally use the SVD dataset~\cite{svddataset}, which provides stereoscopic videos from modern devices (iPhone Pro, Apple Vision Pro). For spatial accuracy, we use two standard metrics: Absolute Relative Error (AbsRel) and Threshold Accuracy ($\delta_1$). For measuring computational performance, we report Frames Per Second~(FPS). While AbsRel and $\delta_1$ calculate spatial accuracy, we propose the Temporal Gradient Square Error~(TGSE) to evaluate temporal stability in dynamic videos containing object motion, as this direct L2 metric robustly penalizes severe flicker artifacts.

\begin{equation}
\nabla_t D^{(t)}(i,j) = D^{(t+1)}(i,j) - D^{(t)}(i,j)
\end{equation}
\begin{equation}
\mathrm{TGSE} = 
\sum_{i,j} 
\big( 
\nabla_t \hat{D}^{(t)}(i,j) - \nabla_t D^{(t)}(i,j)
\big)^2
\end{equation}

\subsection{Experimental Setup}
We conducted experiments in two distinct environments: a high-end setup using \textbf{a single NVIDIA RTX A6000 GPU} for training and benchmarking, and an embedded platform, \textbf{the NVIDIA Jetson Orin Nano (8GB)}, to assess real-world applicability and edge device inference efficiency. This dual-setup ensures our model's capabilities across different hardware tiers. Full implementation details, including model architecture, training parameters, costs, and sampling strategies, are provided in Appendix A.
\subsection{Fast and Stable Video Depth Estimation}
\begin{figure}[h]
\captionsetup{justification=justified, singlelinecheck=false}
\centering
\includegraphics[width=1.05\linewidth]{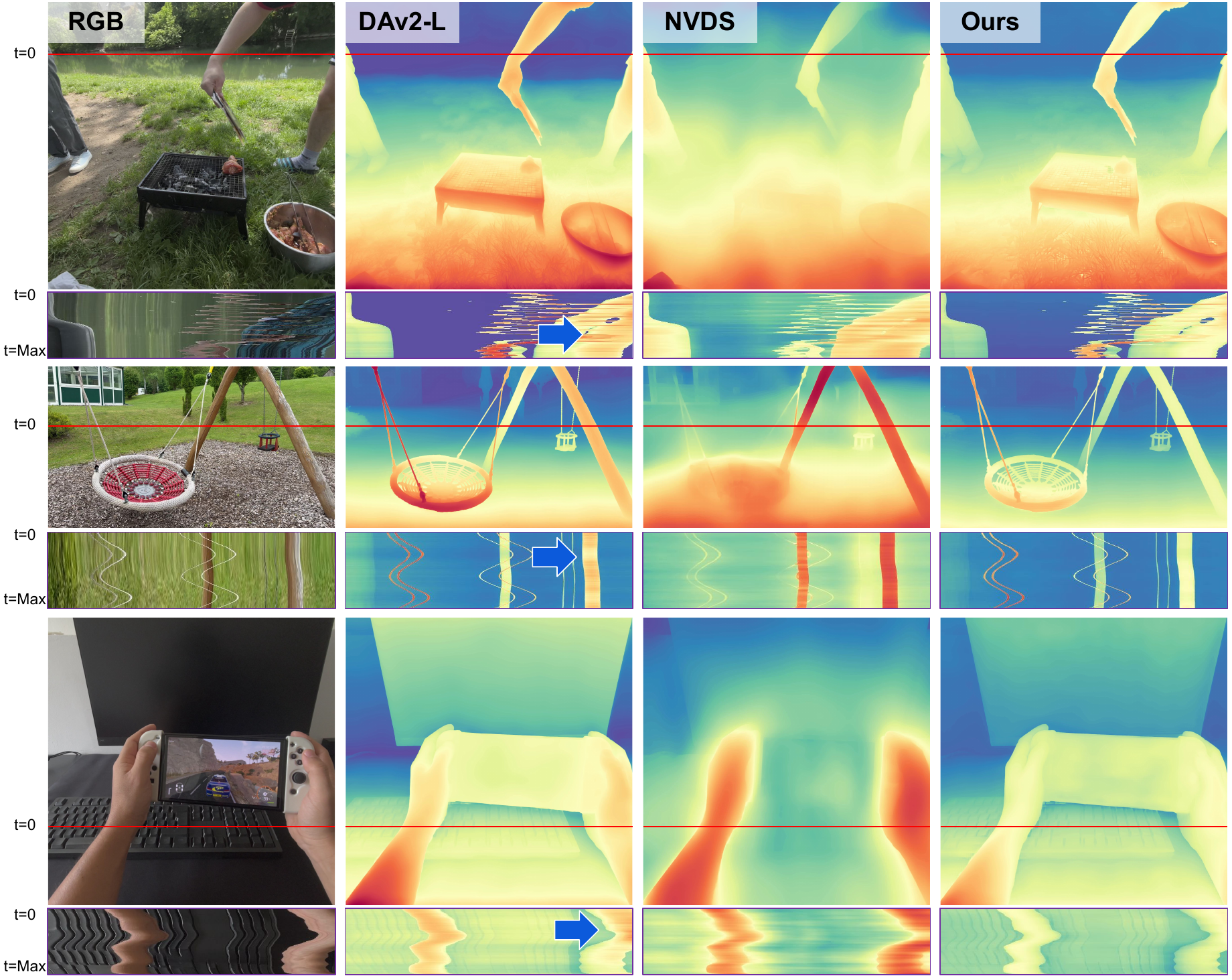}
\caption{\textbf{Quality comparison of depth methods.} We visualize temporal stability using slit-scan image (\textcolor{PurpleBox}{purple box}), generated by stacking a \red{red pixel line} over time. Ideally, static objects (e.g., swinging structure and arms on the desk) should appear as straight horizontal lines. However, DAv2-L exhibits severe temporal flickering (highlighted by \blue{blue arrows}). NVDS struggle with detail depth representation in all videos. Furthermore, VDA-S represents the closer object and the farther object at the similar depth. But VDPP achieve all detail depth representations.}
\label{fig:comparisonDAv2}
\end{figure}

We provide comprehensive quantitative and qualitative comparisons, evaluating VDPP against existing post-precessing methods across diverse benchmarks. We assess performance based on inference speed (FPS) in Table~\ref{tab:speed}, spatial accuracy (AbsRel and $\delta_1$) and temporal consistency (TGSE) in Table~\ref{tab:performance}, on A6000 GPU. Figures~\ref{fig:teaser},~\ref{fig:comparisonDAv2} demonstrate the qualities of VDPP through visual comparisons.

\textbf{Exceptional Speed for Real-Time Performance.}
As shown in Table~\ref{tab:speed}, VDPP with DAv2-B~\cite{depthanythingv2} achieves exceptional speed in post-processing mothods, reaching 47 FPS on NYUv2, 68 FPS on Bonn, and 63 FPS on Sintel—significantly fast while maintaining superior accuracy as shown in Table~\ref{tab:performance}. More importantly, this speed advantage stems from VDPP's efficient architecture: the post-processing module adds only 1.4 to 8.6 ms overhead, enabling \textbf{fast image to depth model} to achieve real-time video depth estimation. Even with the larger DAv2-L backbone, VDPP maintains real-time performance while existing post-processing methods like NVDS suffer from extreme computational overhead, making them impractical for real-world deployment. Notably, VDPP even outperforms the end-to-end model VDA-S and VDA-L~\cite{vda} on several speed benchmarks, while maintaining, particularly in spatial accuracy and temporal consistency.

\textbf{Exceptional Performance: Matching Superior Accuracy with Temporal Consistency.} 
As shown in Table~\ref{tab:performance}, VDPP achieves spatial accuracy and temporal consistency comparable to state-of-the-art end-to-end models across all benchmarks. More importantly, VDPP dramatically outperforms the existing post-processing method NVDS~\cite{nvds} in both spatial accuracy and temporal coherence. Compared to NVDS with the same backbone (DAv2-L), VDPP achieves substantial improvements across all datasets while being up to $34\times$ faster (Table~\ref{tab:speed}). By operating on geometric representations and learning fine-grained residual corrections rather than coarse adjustments, VDPP successfully bridges the accuracy gap between post-processing and end-to-end approaches.

\begin{table*}[htp!]
\centering
\caption{\textbf{Performance comparison on video datasets.}
While end-to-end models achieve high accuracy at the expense of speed and scalability, and existing post-processing methods often suffer from poor spatial accuracy and temporal consistency, our VDPP achieves stable performance. In terms of spatial accuracy (AbsRel, δ₁) and temporal consistency (TGSE), VDPP performs on par with state-of-the-art end-to-end models VDA~\cite{vda}, while significantly outperforming existing post-processing approaches such as NVDS~\cite{nvds}. The numbers in parentheses indicate the relative performance to the baselines (DPT~\cite{dpt}, DAv2~\cite{depthanythingv2}): \textcolor{GoodColor}{Green}/\textcolor{BadColor}{Red} denote improvement/degradation. Among post-processing methods, \textbf{Bold} and \ul{underlined} denote the best and next-best performance, respectively.}
\label{tab:performance}
\footnotesize
\setlength{\tabcolsep}{2pt}
\renewcommand{\theadfont}{\footnotesize}
\sisetup{round-mode=places} 
\resizebox{\textwidth}{!}{ 
\begin{tabular}{l
                S[table-format=1.3, round-precision=3] S[table-format=1.3, round-precision=3] S[table-format=1.3, round-precision=3]
                S[table-format=1.3, round-precision=3] S[table-format=1.3, round-precision=3] S[table-format=1.3, round-precision=3]
                S[table-format=1.3, round-precision=3] S[table-format=1.3, round-precision=3] S[table-format=1.3, round-precision=3]
                }
\toprule
& \multicolumn{3}{c|}{NYUv2~\cite{nyuv2}} & \multicolumn{3}{c|}{Bonn~\cite{bonn}} & \multicolumn{3}{c}{Sintel~\cite{sintel}} \\
\cmidrule(lr){2-4}\cmidrule(lr){5-7}\cmidrule(lr){8-10}
\multirow{-2}{*}{\textbf{Method}}
& {AbsRel$\downarrow$} & {$\delta_1\uparrow$} & {10$^{2}$$\cdot$TGSE$\downarrow$}
& {AbsRel$\downarrow$} & {$\delta_1\uparrow$} & {10$^{5}$$\cdot$TGSE$\downarrow$}
& {AbsRel$\downarrow$} & {$\delta_1\uparrow$} & {10$^{2}$$\cdot$TGSE$\downarrow$} \\
\midrule
\multicolumn{10}{l}{\textit{End-to-end}} \\
VDA-S &0.1118 &0.8926 &1.4338 &0.0428 &0.9885 &4.9792 &0.4517 &0.6266 &1.6015 \\
VDA-L &0.0924 &0.9312 &1.0575 &0.0387 &0.9886 &5.2152 &0.3816 &0.6722 &1.2769 \\
\midrule
\multicolumn{10}{l}{\textit{Image to Depth}} \\
DPT-L &0.2033 &0.6589 &3.2012 &0.0633 &0.9688 &4.8596 &0.6573 &0.5633 &2.7736 \\
DAv2-B &0.1406 &0.8214 &2.0696 &0.0536 &0.9745 &6.0549 &0.4835 &0.6141 &1.9782 \\
DAv2-L &0.1392 &0.8258 &2.0337 &0.0535 &0.9740 &5.5499 &0.4589 &0.6196 &1.7402 \\
\midrule
\multicolumn{10}{l}{\textit{Post-processing}} \\
DPT-L + NVDS &{0.234 \bad{+15.1\%}} &{0.611 \bad{-7.2\%}} &{3.523 \bad{+10.0\%}} &{0.104 \bad{+64.6\%}} &{0.907 \bad{-6.4\%}} &{\ul{4.509} \good{-7.2\%}} &{0.616 \good{-6.3\%}} &{0.538 \bad{-4.6\%}} &{2.543 \good{-8.3\%}} \\
DAv2-B + NVDS &{0.227 \bad{+61.4\%}} &{0.617 \bad{-24.9\%}} &{3.390 \bad{+63.8\%}} &{0.099 \bad{+84.9\%}} &{0.918 \bad{-5.8\%}} &{4.665 \good{-23.0\%}} &{0.512 \bad{+5.9\%}} &{0.572 \bad{-6.9\%}} &{2.360 \bad{+19.3\%}} \\
DAv2-L + NVDS &{0.226 \bad{+62.3\%}} &{0.620 \bad{-25.0\%}} &{3.369 \bad{+65.7\%}} &{0.099 \bad{+85.4\%}} &{0.918 \bad{-5.7\%}} &{4.629 \good{-16.6\%}} &{0.500 \bad{+8.9\%}} &{0.574 \bad{-7.3\%}} &{2.231 \bad{+28.2\%}} \\
\textbf{DPT-L + Ours} &{\ul{0.196} \good{-3.4\%}} &{0.666 \good{+1.1\%}} &{\ul{2.772} \good{-13.4\%}} &{0.061 \good{-4.4\%}} &{\ul{0.974} \good{+0.6\%}} &{\textbf{3.986} \good{-18.0\%}} &{0.588 \good{-10.5\%}} &{0.580 \good{+2.9\%}} &{2.342 \good{-15.6\%}} \\
\textbf{DAv2-B + Ours} &{\textbf{0.132} \good{-6.0\%}} &{\ul{0.844} \good{+2.8\%}} &{\textbf{1.789} \good{-13.6\%}} &{\ul{0.044} \good{-18.8\%}} &{\textbf{0.986} \good{+1.2\%}} &{5.785 \good{-4.5\%}} &{\ul{0.445} \good{-8.0\%}} &{\ul{0.653} \good{+6.4\%}} &{\ul{1.753} \good{-11.4\%}} \\
\textbf{DAv2-L + Ours} &{\textbf{0.132} \good{-5.2\%}} &{\textbf{0.845} \good{+2.4\%}} &{\textbf{1.789} \good{-12.0\%}} &{\textbf{0.042} \good{-20.7\%}} &{\textbf{0.986} \good{+1.2\%}} &{5.636 \bad{+1.6\%}} &{\textbf{0.412} \good{-10.1\%}} &{\textbf{0.656} \good{+5.8\%}} &{\textbf{1.563} \good{-10.2\%}} \\
\bottomrule
\end{tabular}
} 
\end{table*}

\textbf{Visual Quality with High Spatial Detail and Temporal Stability}.
Beyond the quantitative results, the qualitative results in Figure~\ref{fig:teaser} and ~\ref{fig:comparisonDAv2} visually confirm VDPP's outstanding handling of both spatial detail and temporal stability. These results were generated from a zero-shot experiment on the SVD dataset~\cite{svddataset}, using DAv2-L as the I2D model. In Figure~\ref{fig:comparisonDAv2}, static objects such as a person's stationary torso whose depth should remain stable regardless of hand motion, or the swinging structure, should ideally appear as straight, horizontal lines. However, the baseline DAv2-L shows severe temporal flickering, visible as distorted and broken lines(blue arrow). NVDS struggle to represent fine details, collapsing near and far objects into a similar depth range. In sharp contrast, VDPP produces clean, stable horizontal lines, demonstrating high temporal coherence, while simultaneously preserving intricate spatial details and accurate depth separation in the scene. Ultimately, VDPP avoids the flickering of frame-by-frame methods and the depth ambiguity of other models, producing temporally coherent and detailed depth sequences across diverse scenarios. Due to space constraints, additional qualitative comparisons including both post-processing methods and E2E models such as VDA~\cite{vda} are provided in Appendix B.

\textbf{Performance in Small Resolution.}
Resource-constrained environments often require the use of lower input resolutions. To validate robustness in such scenarios and across different input scales, we additionally evaluate all methods by center-cropping to a 384×384 resolution on the Sintel dataset~\cite{sintel}. VDPP maintains its superior performance across all three dimensions: achieving SOTA-level speed while preserving both spatial fidelity and temporal coherence. These results confirm that our geometric downsampling approach scales effectively across various deployment scenarios, from high-resolution applications to resource-constrained environments. The detailed results are provided in Appendix C.

\begin{figure}[b!]
\centering
\includegraphics[width=0.8\linewidth]{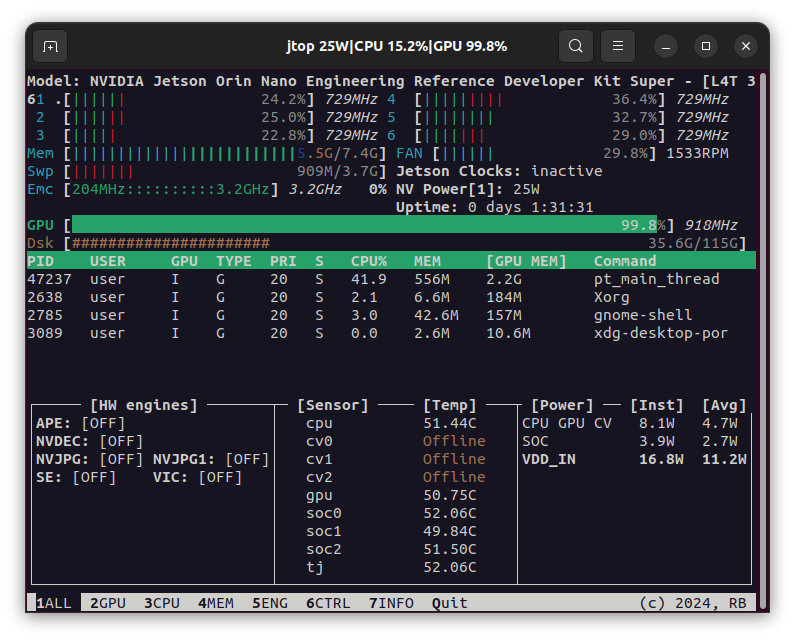}
\caption{VDPP running on NVIDIA Jetson Orin Nano (8GB) at 43.5 FPS with 5.5 GB memory usage. Other temporal methods (NVDS, VDA-L, VDA-S) failed to run on this edge device due to memory overflow as shown in Appendix E, validating VDPP's practical deployment advantage.}
\label{fig:jetson_success}
\end{figure}

\subsection{Memory Efficiency on Lightweight Devices}

The significant advantage of VDPP in computational efficiency extends crucially to memory usage, enabling deployment on resource constrained devices. While powerful attention mechanisms are essential for temporal coherence, their memory requirements scale quadratically with the input resolution ($(H \times W)^2$). This creates a fundamental bottleneck: VDA~\cite{vda} processes high-resolution RGB inputs directly, leading to prohibitively large attention maps. In contrast, VDPP's Efficient Geometric Refinement paradigm is the key enabler on significantly lower-resolution representations without sacrificing quality. We validated this advantage experimentally on an \textbf{NVIDIA Jetson Orin Nano (8GB)}, a representative edge device. \textbf{On this platform, our VDPP models ran successfully}, achieving 43.5 FPS while consuming $\approx$5.5 GB of memory as shown in Figure~\ref{fig:jetson_success} and Appendix E. As expected, the image-depth base model (DAv2~\cite{depthanythingv2}) also ran successfully. In contrast, all competing temporal methods such as NVDS, VDA-L, and even VDA-S failed completely, either exceeding 7 GB memory usage or causing system crashes due to Out-Of-Memory errors. Notably, when we disabled the downsampling component in VDPP for ablation, it also failed to run, confirming that our downsampling-based paradigm is not just an optimization, but a necessity for practical deployment. VDPP achieves this edge deployment without additional model compressions such as distillation or quantization.

\subsection{Scalability to Different Backbones}

As the flexibility of depth-only post-processing paradigm, VDPP can seamlessly integrate with any image depth estimator, enabling users to optimize for their requirements—whether prioritizing speed for real-time applications or accuracy for high-fidelity reconstruction. This plug-and-play capability future-proofs the framework: as better image depth estimation models emerge, they can immediately benefit from VDPP without retraining the backbone. Table~\ref{tab:speed} and ~\ref{tab:performance} demonstrate VDPP's compatibility with diverse image depth estimators (DPT-L, DAv2-Base, DAv2-Large). This flexibility allows practitioners to select the configuration for systems demanding the highest quality.

\begin{figure}[!t]
\captionsetup{justification=justified, singlelinecheck=false}
\centering
\includegraphics[width=\linewidth]{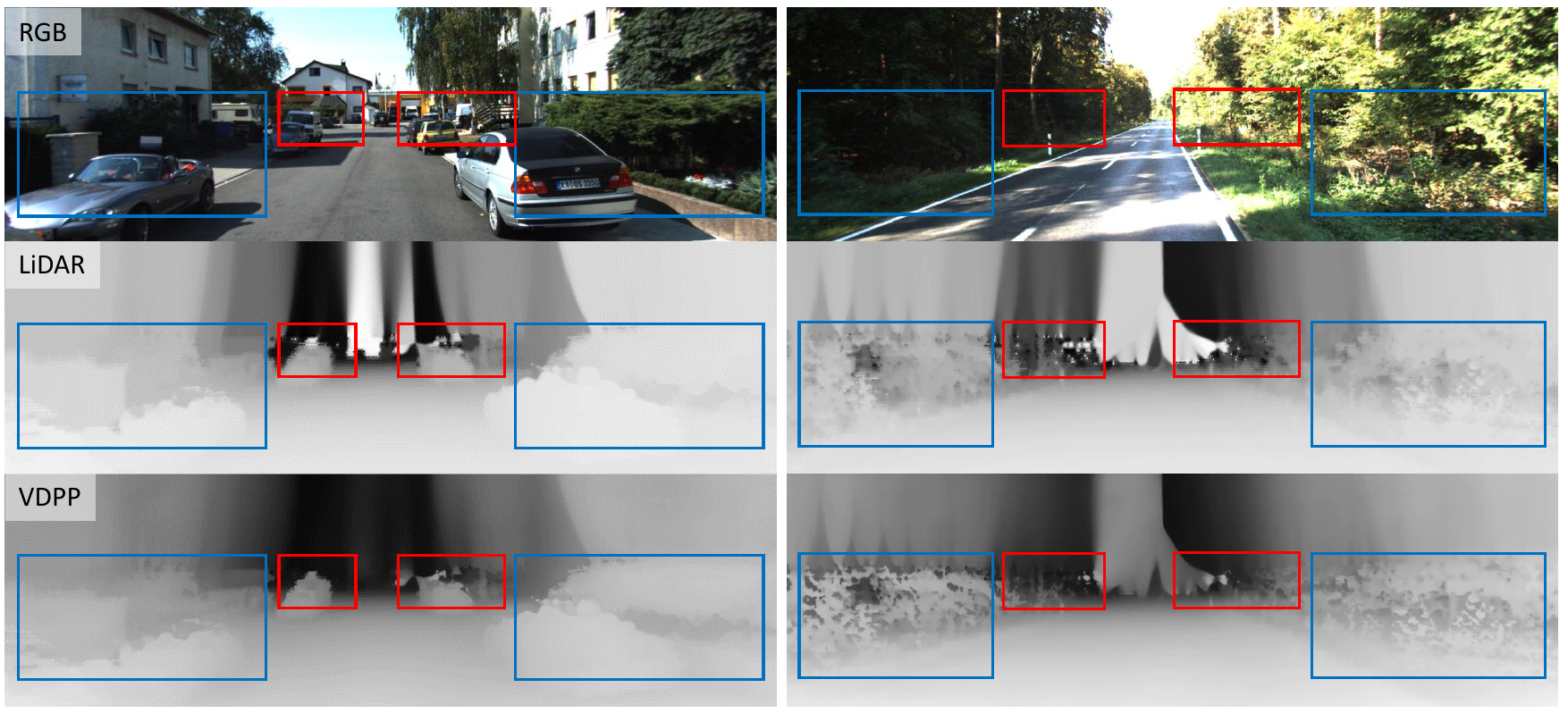}
\caption{\textbf{Post-processing to LiDAR dataset. VDPP enhances LiDAR-derived depth quality on KITTI dataset~\cite{kitti}}. Raw LiDAR point clouds are inpainted to generate dense depth maps, which are then refined by VDPP. Red boxes highlight the reduction of horizontal artifacts caused by the LiDAR scanning method. Blue boxes demonstrate significantly improved spatial detail depth representation, showcasing VDPP's effectiveness as a sensor-agnostic post-processing solution.}
\label{fig:lidar}
\end{figure}

\subsection{Application to Depth Sensors without RGB}
A crucial advantage of VDPP's depth-only architecture is its direct applicability to real depth sensors such as LiDAR and ToF. Unlike RGB-dependent methods, VDPP operates directly on depth maps, a capability critical for applications such as autonomous driving and robotics. We validated this on the KITTI dataset~\cite{kitti}, where raw LiDAR point clouds are typically processed through depth completion (inpainting sparse points) to generate dense depth maps. However, as noted by~\cite{depthanythingv2}, even depth data measured by hardware devices can be noisy and imperfect. Consequently, these completed depths often suffer from spatial and temporal inconsistencies due to moving objects and sensor noise. Figure~\ref{fig:lidar} demonstrates VDPP's effectiveness on real sensor data. The raw inpainted LiDAR exhibit inconsistencies around object boundaries and distant regions where point cloud density is low. After VDPP refinement, these inconsistencies are significantly reduced while preserving spatial accuracy and smoothing horizontal artifacts. Additional results on LiDAR processing are provided in Appendix D.

\subsection{Ablation Study}
To verify that each design choice in VDPP is essential, we conducted a comprehensive ablation study on the Sintel dataset~\cite{sintel} using depth maps from DAv2-L~\cite{depthanythingv2} as input to our model, with results summarized in Table~\ref{tab:ablation}. 


\vspace{-3mm}
\paragraph{Differential Scaler}
The median-based Differential Scaler is crucial for spatial accuracy. As shown in Table~\ref{tab:ablation}, removing this module (`w/o Diff. Scaler') significantly degrades performance. In contrast, replacing our robust median scaler with a simple `Average Pooling' worsened AbsRel, confirming that robust statistics are essential for handling depth ambiguity. While average pooling offers an improvement in speed, this scaling method degrades spatial accuracy more than our differential scaler.
\vspace{-3mm}
\paragraph{Spatial and Temporal Refinement}
Both Surface Normals and Residual Prediction contribute to the delicate improvement of spatial and temporal quality. Although ‘w/o Surface Normal’ shows a marginal improvement in $\delta_1$ as an optimization trade-off, removing either component disrupts the full model's spatial-temporal balance, leading to a clear degradation in temporal consistency (TGSE). Residual prediction, in particular, offers a significant benefit to temporal consistency.

\begin{table}[t]
\centering
\caption{We conducted an ablation study on the Sintel dataset~\cite{sintel} to validate the contributions of each component in addressing the trade-off between speed and fidelity. The table illustrates the performance degradation upon each is removal, confirming their individual importance. \textbf{Bold numbers} indicate the best performance, and \ul{underlined numbers} represent the second-best.}
\label{tab:ablation}
\small
\setlength{\tabcolsep}{6pt}
\begin{tabular}{l|cc|c}
\toprule
\multirow{2}{*}{\textbf{Method}} & \multicolumn{2}{c|}{\textbf{Spatial}} & \textbf{Temporal} \\
 & AbsRel↓ & $\delta_1$↑ & 10$^{2}$$\cdot$TGSE↓ \\
\midrule
\textbf{Full Model} &\textbf{0.4142} &\ul{0.6464} &\textbf{1.5616} \\
\midrule
w/o Diff. Scaler  &\ul{0.4276} &0.6335 &1.5667 \\
w/ Average Pooling &0.4325 &0.6463 &\ul{1.5623} \\
\midrule
w/o Surface Normal &0.4431 &\textbf{0.6476} &1.6500 \\
w/o Residual Pred. &0.4375 &0.6443 &1.8042 \\
\bottomrule
\end{tabular}
\end{table}


\subsection{Limitation} 
A primary limitation of our study lies in the restricted number of post-processing baselines available for comparison. While there is an abundance of end-to-end video depth models, plug-and-play post-processing methods has been comparatively sparse. As a result, our comparative evaluation primarily focuses on NVDS, which stands as the only accessible latest baseline in post-processing. However, the practical advantages of post-processing methods remain compelling for real-world deployment. we release VDPP as an open-source baseline to encourage efficient research into depth-only temporal refinement.

\section{Conclusion}

We presented VDPP, a video depth post-processing framework that improves the post-processing approach to video depth estimation. By re-framing the task from a computationally heavy, RGB-dependent scene reconstruction into a targeted geometric refinement problem, VDPP successfully bridges the gap between the flexibility of post-processing and the performance of monolithic end-to-end models.

Our framework's efficacy stems from operating purely on geometric representations. By combining efficient geometric downsampling with residual learning in the error space, VDPP minimizes computational overhead, achieving real-time inference speed that significantly outperforms existing post-processing baselines while maintaining spatial fidelity and temporal coherence. This design yields three practical advantages. First, it enables real-time deployment on memory-constrained edge devices, such as the NVIDIA Jetson Orin Nano, which previously struggled with video depth models. Second, by eliminating RGB dependency, it ensures true plug-and-play scalability, resolving the adaptation lag of monolithic models and allowing immediate integration of emerging single-image depth estimators or depth-only sensors without full-scale retraining. Third, the lightweight geometric refinement architecture matches the temporal coherence of heavy E2E systems at a fraction of the computational cost. Ultimately, as the field continues to advance with more powerful image depth estimators, VDPP provides a sustainable and practical path forward for robust real-time 3D scene understanding.

\section*{Acknowledgment}
This work was supported by the Korean Government through the grants from IITP (RS-2021-II211343, RS-2025-25442338, 26-InnoCORE-01).

\clearpage
{
    \small
    \bibliographystyle{ieeenat_fullname}
    \bibliography{main}
    \clearpage
}
\begin{strip}
\centering
\LARGE\textbf{VDPP: Video Depth Post-Processing \\for Speed and Scalability}\\[2pt]
{\Large Supplementary Material}
\end{strip}

\appendix

\begin{figure*}[t]
\captionsetup[figure]{justification=justified, singlelinecheck=false}
\centering
\includegraphics[width=\linewidth]{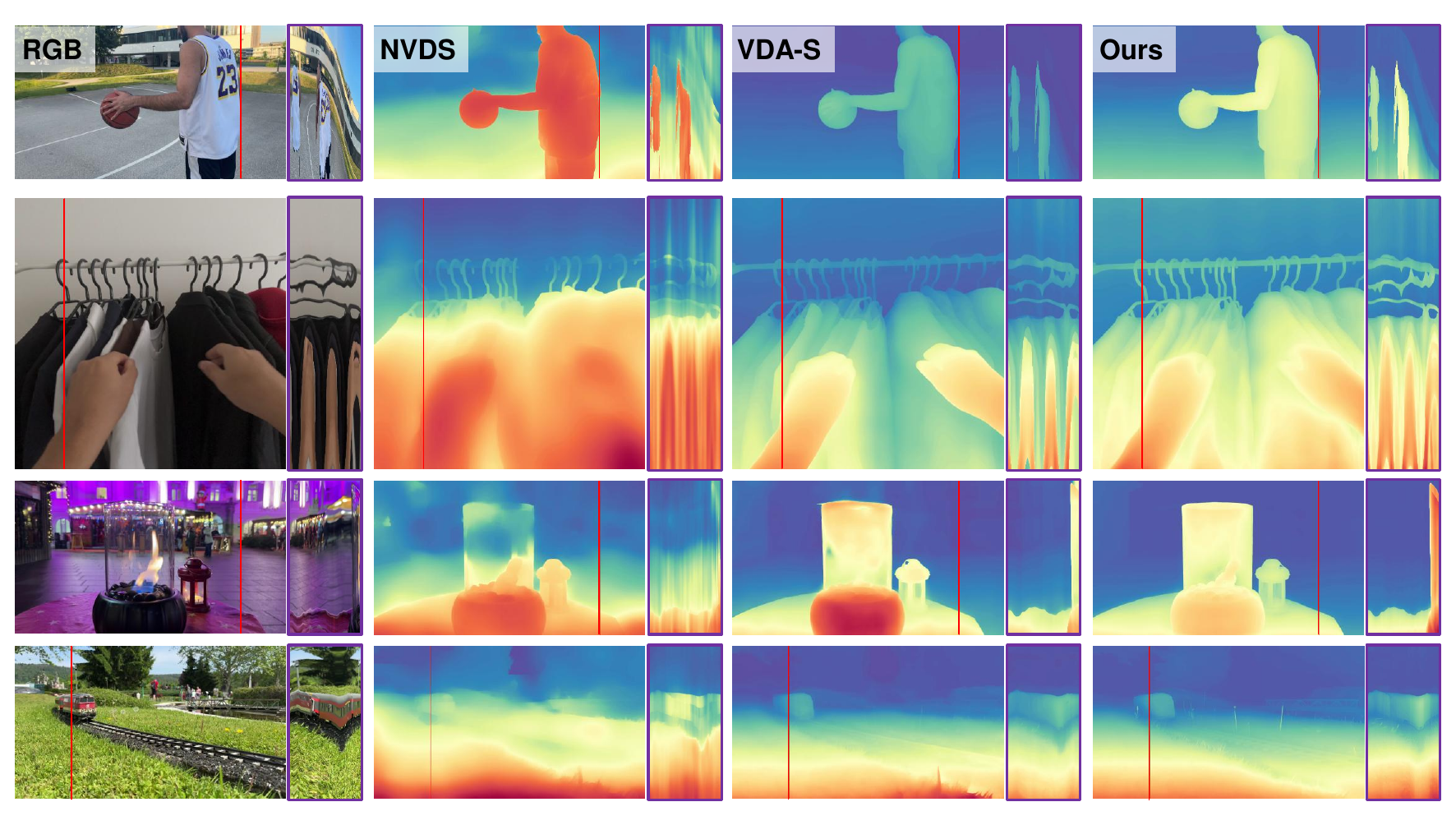}
\caption{
    \textbf{Qualitative comparison of VDPP with NVDS (post-processing) and VDA (end-to-end).}
    We compare video depth estimation by accumulating results along the time axis. The temporal profile (\textcolor{PurpleBox}{purple box}) is a slit-scan image generated by stacking the \red{red pixel line} from each frame sequentially over time. (Top row) A basketball player runs forward and then moves away. Both \textbf{NVDS} and \textbf{VDA} show minimal depth change, whereas \textbf{VDPP} correctly captures the motion. (Second row) A hand moves through clothes on a hanger. The slit-scan image reveals \textbf{NVDS} suffers from severe \textbf{flickering}, even in areas that should be static. \textbf{VDA} lacks fine-grained depth expression. (Third row) Flames flicker inside a transparent fire pit. \textbf{NVDS} again exhibits heavy flickering, and \textbf{VDA} fails to distinguish the depth between the front and back of the pit. \textbf{VDPP} clearly shows the depth difference with almost no flickering. (Bottom row) A train moves laterally across tracks. \textbf{NVDS} fails to estimate depth during the fast motion, while \textbf{VDA} loses the details of the tracks. In all scenarios, \textbf{VDPP} demonstrates superior \textbf{spatial accuracy} and \textbf{temporal coherence}.
}
\label{fig:app_comparison}
\end{figure*}

\begin{figure*}[t]
\captionsetup[figure]{justification=justified, singlelinecheck=false}
\centering
\includegraphics[width=\linewidth]{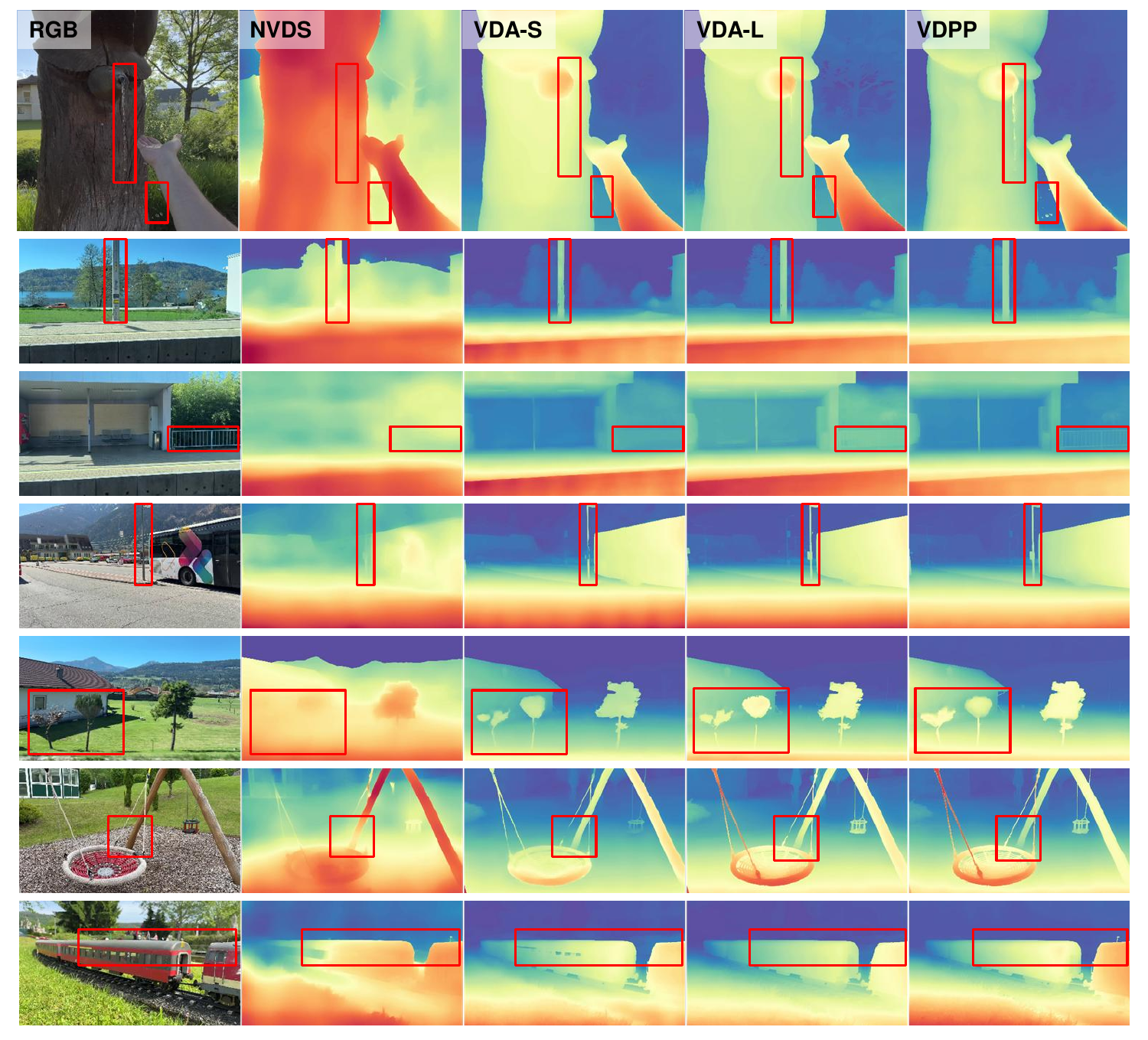}
\caption{
    \textbf{Qualitative comparison of fine-grained spatial fidelity across seven challenging scenarios.}
    From left to right: RGB Input, NVDS, VDA-Small, VDA-Large (SOTA), and our VDPP.
    \textbf{(Top row)} The \textit{Water falling} scene tests the ability to resolve fast-moving, non-rigid water and individual droplets. \textbf{(Second row)} The \textit{Power Lines} scene highlights the challenge of preserving thin, static structures. \textbf{(Third row)} The \textit{Fence} scene evaluates the separation of fine foreground posts from a complex background.
    \textbf{(Fourth row)} The \textit{Bus Stop} tests robustness to dynamic background motion (passing buses) behind a thin object. \textbf{(Fifth row)} The \textit{Street Trees} scene challenges the model to maintain structural integrity during complex camera and background motion.
    \textbf{(Sixth row)} The \textit{Swing} scene focuses on resolving intricate details like thin ropes against a main structure. \textbf{(Bottom row)} The \textit{Train} scene tests the difficult combination of transparent surfaces (windows) and fine details (antennas). In all examples, VDPP demonstrates a superior ability to preserve geometric integrity where other methods introduce artifacts, over-smoothing, or structural breaks.
}
\label{fig:app_finedetail}
\end{figure*}

{
\Large
\part*{Appendix}
\vspace{-3mm}
}
In this supplementary material, we provide additional implementation details in Appendix~\ref{sec:impl_details}. We then offer an in-depth qualitative analysis (Appendix~\ref{app:quality}) against competing methods, visually validating VDPP's superior spatio-temporal coherence in Figure~\ref{fig:app_comparison},~\ref{fig:app_finedetail}. Furthermore, we present critical evidence of practical viability, starting with comprehensive performance metrics on limited resolutions in Table~\ref{tab:384} (Appendix~\ref{app:small_resol}). We follow this with dedicated validation on real-world LiDAR sensor data in Figure~\ref{fig:app_LiDar} (Appendix~\ref{app:lidar}), and conclude with a decisive memory efficiency profiling on the NVIDIA Jetson Orin Nano in Figure~\ref{fig:jetson_success} and Figure~\ref{fig:jetson_fail} (Appendix~\ref{app:jetson}). 
Finally, we validate the reliability of TGSE as a temporal consistency metric through sensitivity analysis in Figure~\ref{fig:tgse} (Appendix~\ref{app:tgse}).

\vspace{2em}
\section{Implementation Details}
\label{sec:impl_details}
This section provides the comprehensive implementation details for our VDPP framework, covering the datasets, data processing, architectural choices, and training hyperparameters.
\paragraph{Training dataset details.}
For video training, we utilize three datasets with precise depth annotations: VKITTI~\cite{vkitti}, TartanAir~\cite{tartanair}, and PointOdyssey~\cite{pointodyssey}.
Initially, we employ an Image-to-Depth model~\cite{depthanythingv2} to generate depth estimations from the full-resolution datasets. Our post-processing network is then trained using these generated depth maps.
For training, the datasets are processed into non-overlapping, consecutive 16-frame sequences. We apply random cropping to obtain 224x224 patches, which serve as the input resolution.
To address the uneven data distribution and prevent domination by large-scale datasets (e.g., TartanAir), we employ a Weighted Random Sampler for uniform sampling.

\paragraph{Implementation details.}Our model consists of several components. For the encoder, we utilized the small version of DINOv2 encoder~\cite{dinov2} and trained it from scratch. For the decoder, we adopted and fine-tuned the small version head from Video Depth Anything~\cite{vda}. All other modules were also trained from scratch. We train the model using the AdamW optimizer~\cite{adamw} with a base initial learning rate of 1e-6. We employ a CosineAnnealingWarmRestarts scheduler~\cite{cosineannealing}, configured with $T_0 = 10,000$, $T_{mult} = 2$, and $\eta_{min} = 1e-9$. The batch size is set to 16. The loss weights for $\alpha$ and $\beta$ are set to 1.0 and 10.0, respectively. For downsampling, all depth maps are downsampled using bilinear interpolation with a ratio of $r=0.5$. The model was trained for 430 hours on a single NVIDIA A6000 GPU, for a total of 630K iterations.

\begin{table*}[t]
\centering
\caption{\textbf{Performance and Speed Comparison on Sintel (384x384)}. This table combines speed (FPS, TPF) and performance (AbsRel, $\delta_1$, TGSE) metrics. I2D refers to Image to Depth conversion time, and D2V refers to Depth to Video conversion time. The numbers in parentheses indicate the relative performance compared to the baselines(DPT, DAv2): \textcolor{GoodColor}{green} denotes improvement, and \textcolor{BadColor}{red} denotes degradation. Among post-processing methods, \textbf{Bold} and \ul{underlined} denote the best and next-best performance, respectively.}
\label{tab:384}
\setlength{\tabcolsep}{5pt} 
\begin{tabular}{@{}lrrrrccc@{}}
\toprule 
\multirow{2}{*}{\textbf{Method}} & \multicolumn{4}{c|}{Speed} & \multicolumn{3}{c}{Performance Metrics} \\ \cmidrule(l){2-8} 
 & \multicolumn{1}{c}{FPS ↑} & \multicolumn{1}{c}{Total (ms)} & \multicolumn{1}{c}{I2D (ms)} & \multicolumn{1}{c}{D2V (ms)} & \multicolumn{1}{c}{AbsRel↓} & \multicolumn{1}{c}{$\delta_1\uparrow$} & \multicolumn{1}{c}{10$^{2}$$\cdot$TGSE$\downarrow$} \\ \midrule
\textit{End-to-end} & \multicolumn{1}{l}{} & \multicolumn{1}{l}{} & \multicolumn{1}{l}{} & \multicolumn{1}{l}{} & \multicolumn{1}{l}{} & \multicolumn{1}{l}{} & \multicolumn{1}{l}{} \\
VDA-S & 65.6 & 15.3 & {--} & {--} & 0.355 & 0.681 & 1.760 \\
VDA-L & 18.0 & 55.5 & {--} & {--} & 0.324 & 0.729 & 1.424 \\ \midrule
\textit{Image to Depth} & \multicolumn{1}{l}{} & \multicolumn{1}{l}{} & \multicolumn{1}{l}{} & \multicolumn{1}{l}{} & \multicolumn{1}{l}{} & \multicolumn{1}{l}{} & \multicolumn{1}{l}{} \\
DPT-L & 129 & 7.7 & 7.7 & {--} & 0.548 & 0.610 & 2.545 \\
DAv2-B & 141 & 7.1 & 7.1 & {--} & 0.358 & 0.678 & 1.809 \\
DAv2-L & 75 & 13.3 & 13.3 & {--} & 0.352 & 0.671 & 1.711 \\ \midrule
\textit{Post-processing} & \multicolumn{1}{l}{} & \multicolumn{1}{l}{} & \multicolumn{1}{l}{} & \multicolumn{1}{l}{} & \multicolumn{1}{l}{} & \multicolumn{1}{l}{} & \multicolumn{1}{l}{} \\
DPT-L + NVDS & 4 & 225.0 & 7.7 & 217.2 & 0.471 \good{-14.0\%} & 0.615 \good{+0.8\%} & 2.407 \good{ -5.4\%} \\
DAv2-B + NVDS & 4 & 224.3 & 7.1 & 217.2 & 0.387 \bad{+8.1\%} & 0.668 \bad{-1.5\%} & 2.304 \bad{+27.4\%} \\
DAv2-L + NVDS & 4 & 230.5 & 13.3 & 217.2 & 0.384 \bad{+9.1\%} & 0.674 \good{+0.5\%} & 2.180 \bad{+27.4\%} \\
\textbf{DPT-L + Ours} & \multicolumn{1}{r}{{\ul{104}}} & \multicolumn{1}{r}{{\ul{9.6}}} & \multicolumn{1}{r}{7.7} & \multicolumn{1}{r}{1.8} & \multicolumn{1}{r}{0.559 \bad{+2.0\%}} & \multicolumn{1}{r}{0.633 \good{+3.8\%}} & \multicolumn{1}{r}{2.207 \good{-13.3\%}} \\
\textbf{DAv2-B + Ours} & \multicolumn{1}{r}{\textbf{112}} & \multicolumn{1}{r}{\textbf{8.9}} & \multicolumn{1}{r}{7.1} & \multicolumn{1}{r}{1.8} & \multicolumn{1}{r}{\textbf{0.312} \good{-12.9\%}} & \multicolumn{1}{r}{\textbf{0.731} \good{+7.8\%}} & \multicolumn{1}{r}{{\ul{1.756} \good{-3.0\%}}} \\
\textbf{DAv2-L + Ours} & \multicolumn{1}{r}{66} & \multicolumn{1}{r}{15.1} & \multicolumn{1}{r}{13.3} & \multicolumn{1}{r}{1.8} & \multicolumn{1}{r}{{\ul{0.319} \good{-9.3\%}}} & \multicolumn{1}{r}{{\ul{0.713} \good{+6.4\%}}} & \multicolumn{1}{r}{\textbf{1.704} \good{-0.4\%}} \\ \bottomrule
\end{tabular}
\end{table*}

\section{Quality Comparison with Depth Estimation Models Across All Metrics}
\label{app:quality}
In addition to the quantitative metrics presented in the main paper, this section provides an in-depth qualitative analysis to complement our findings in Figure~\ref{fig:app_comparison} and Figure~\ref{fig:app_finedetail}. We visually substantiate our core claim: VDPP resolves the spatial detail and temporal stability, overcoming the failure modes of both post-processing and end-to-end solutions.

\subsection{Spatio-Temporal Coherence Analysis}
The visualizations in Figure \ref{fig:app_comparison} are impressive. They are specifically chosen to represent common, yet challenging real-world scenarios where existing methods typically fail, forcing a compromise between spatial detail and temporal stability. We compare VDPP against NVDS~\cite{nvds} (post-processing) and VDA~\cite{vda} (end-to-end) to visually substantiate our core claim: VDPP resolves this trade-off.

The first two rows (Basketball, Hanger) demonstrate the fundamental conflict between high-frequency detail and stability. The bottom two rows (Fire Pit, Train) test robustness in complex environments with transparency and fast motion.
\begin{itemize} 
    \item \textbf{In the Basketball scene,} the key challenge is capture in large-scale motion (change in depth-of-field). Both NVDS and VDA-S produce an overly `flat' and temporally insensitive result in the slit-scan image (\textcolor{PurpleBox}{purple box}), failing to register the player's significant movement away from the camera. VDPP, however, clearly captures this dynamic depth progression.
    \item \textbf{In the Hanger scene,} the slit-scan image (\textcolor{PurpleBox}{purple box}) provides a definitive diagnosis. NVDS, a post-processing method, reveals its critical flaw: it introduces chaotic, high-frequency oscillations, resulting in severe temporal flickering even in static background regions. Conversely, VDA-S achieves temporal smoothness, but does so by sacrificing spatial fidelity. Its end-to-end approach appears to over-smooth the scene, blurring the individual clothing items into a single mass. VDPP uniquely achieves both goals: a stable temporal profile (a straight line in the static regions) combined with sharp, distinct spatial details.
    \item \textbf{In the Fire Pit scene,} the challenge is twofold: transparency and stochastic (non-rigid) motion from the flames. NVDS breaks down into uncontrolled flickering, unable to handle the non-rigid motion. VDA-S struggles with depth ambiguity, collapsing the depth between the front and back of the transparent pit into a single ambiguous plane. VDPP is the only method to correctly distinguish the distinct foreground and background depth layers while maintaining coherence.
    \item \textbf{In the Train scene,} the model is tested against fast lateral motion. NVDS's depth map almost completely disintegrates, demonstrating an inability to process the rapid pixel displacement. VDA-S maintains temporal stability but at a significant cost: it loses the critical geometric integrity of the tracks, blurring them into a non-descript surface. VDPP again excels, preserving the sharp spatial details of the individual tracks without introducing temporal artifacts or motion blur.
\end{itemize}
These visual results collectively serve as strong qualitative evidence. They confirm that VDPP is not making a simple compromise, but is fundamentally resolving the conflict between spatial accuracy and temporal coherence.

\subsection{Spatial Fidelity and Fine Detail Analysis}
While the previous section focused on temporal stability, this section provides a detailed analysis of spatial fidelity. End-to-end models like VDA often achieve temporal smoothness by over-smoothing spatial details, while post-processing methods like NVDS can fail to resolve fine structures. Figure~\ref{fig:app_finedetail} compares VDPP against NVDS, VDA-S, and the SOTA VDA-L in seven challenging scenarios focused on fine-grained detail.

\begin{itemize}
    \item \textbf{In the Waterfall scene,} the challenge is capturing fast-moving, non-rigid water. Most methods fail to represent the stream's depth. Even the SOTA VDA-L, while capturing the main stream, fails to resolve the individual falling droplets. VDPP is the only method to clearly represent both the continuous stream and the distinct, fast-moving droplets, demonstrating its superior depth expression.
    \item \textbf{In the Power Lines Structure scene,} when zoomed in, VDPP is the only method to render a clean, complete power pole. NVDS fails to identify the structure, while both VDA-S and VDA-L produce results with ``holes'' where depth is missing, revealing instability even in static structures. VDPP maintains both spatial integrity and temporal stability.
    \item \textbf{In the Fence scene,} against a complex background of dense trees, VDPP is the only method to accurately capture the fine geometry of the individual fence posts and the gaps between them. All competitors, including VDA-L, render the general shape of the fence but completely fail to resolve these fine-grained structural details.
    \item \textbf{In the Bus Stop Sign scene,} the challenge is a static thin pole against a dynamic background of passing buses. This causes both NVDS and VDA-S to incorrectly blend the pole's depth with the background in several frames. VDA-L avoids blending but introduces significant depth artifacts not present in the original scene. VDPP remains robust, producing a clean and accurate representation of the sign throughout the sequence.
    \item \textbf{In the Street Trees scene,} viewed from a moving bus, the complex and dynamic parallax causes NVDS, VDA-S, and even VDA-L to fail. As a house passes in the background, their representations of the smaller tree trunks ``break'' or become disconnected. VDPP is the only method to handle this challenging dynamic scene, correctly preserving the geometric integrity of the trees.
    \item \textbf{In the Swing scene,} All competing methods, including VDA-L, fail to distinguish the thin ropes of the swing from its main structure, blending them into a single object. VDPP successfully preserves the depth separation and resolves these fine structures.
    \item \textbf{In the Train scene,} NVDS and VDA-S fail a classic challenge: transparent windows. They are unable to estimate depth for the window, likely due to the visible background. VDA-L correctly handles the transparent window but fails to capture finer details like the train's antenna. VDPP successfully resolves both the complex transparent surface and the fine-grained antenna detail.
\end{itemize}

These results confirm that VDPP's refinement process excels at preserving and enhancing spatial fidelity, overcoming the over-smoothing tendencies of VDA and the structural failures of NVDS.

\begin{figure*}[t]
\centering
\includegraphics[width=\linewidth]{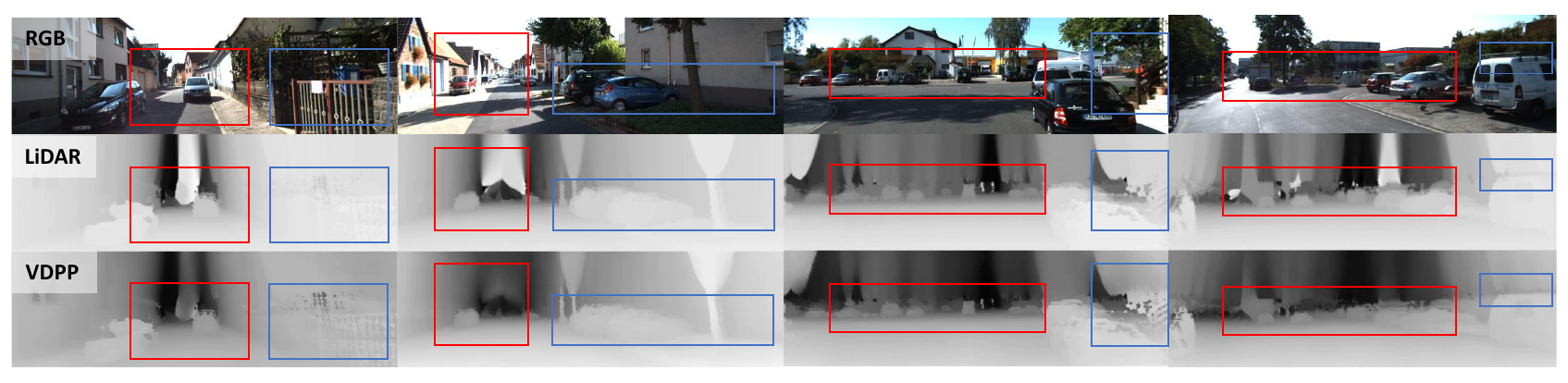}
\caption{\textbf{Qualitative validation of VDPP on real-world LiDAR data from the KITTI dataset~\cite{kitti}.} We compare the ``Raw Inpainted'' depth (from LiDAR depth completion) against our refined ``VDPP'' output across four driving sequences. The \textcolor{red}{red boxes} highlight VDPP's ability to \textbf{suppress artifacts} inherent to the inpainting process, such as horizontal streaking. The \textcolor{blue}{blue boxes} demonstrate how VDPP \textbf{enhances depth expression}, clarifying object boundaries and improving geometric detail where the raw input was poorly defined. This validation confirms VDPP is not limited to refining single-image models but also serves as a powerful plug-and-play post-processor, significantly improving depth quality from real-world 3D sensor pipelines.}
\label{fig:app_LiDar}
\end{figure*}

\section{Comparison on Small Resolution}
\label{app:small_resol}
To further validate VDPP's practical viability, we conducted a comprehensive benchmark at a fixed 384x384 resolution on the Sintel dataset~\cite{sintel}. This setup mimics common industrial applications where computational and memory budgets are strictly limited, forcing the use of downscaled inputs. The results, detailed in Table \ref{tab:384}, decisively demonstrate VDPP's ability to resolve the performance trilemma, led by its state-of-the-art speed.

\textbf{State-of-the-Art Speed:} The primary advantage of VDPP is its exceptional processing speed. As shown in Table \ref{tab:384}, our DAv2-B + Ours configuration achieves an extremely high 112 FPS. This is massively faster than all end-to-end competitors, running approximately 6.2x faster than VDA-L and 1.7x faster than VDA-S. Even our DAv2-L + Ours setup (66 FPS) surpasses VDA-S. This speed is enabled by our highly efficient D2V (Depth-to-Video) module, which adds negligible overhead.

\textbf{Superioir Spatial Accuracy Despite High Speed:} Crucially, this SOTA-level speed is not a trade-off. VDPP achieves this while simultaneously delivering state-of-the-art spatial accuracy. Both our DAv2-B + Ours (0.312 AbsRel) and DAv2-L + Ours (0.319 AbsRel) configurations are more accurate than the heavyweight VDA-L (0.324 AbsRel). This proves VDPP not only stabilizes the base models but significantly refines their geometric quality, a feat the competing post-processing method (NVDS) fails, as it severely degrades accuracy.

\textbf{Competitive Temporal Coherence:} Furthermore, this unprecedented combination of speed and accuracy is achieved without sacrificing temporal stability. Our temporal coherence (TGSE) scores are better than VDA-S and remain highly competitive with the VDA-L baseline. This confirms that our framework's gains in speed and accuracy do not introduce the flickering artifacts that plague other fast methods.

In summary, this 384x384 experiment confirms VDPP's exceptional value for practical deployment. It is the only framework that delivers SOTA speed, SOTA spatial accuracy, and competitive temporal coherence in a single, lightweight package, confirming its suitability for real-world industrial applications.

\section{Experiment for Real Depth Sensors}
\label{app:lidar}
As introduced in the main paper, one of VDPP's most significant advantages is its sensor-agnostic, depth-only architecture. This allows it to directly refine data from real-world depth sensors, a critical capability for robotics and autonomous systems that RGB-dependent methods lack. This section provides a more detailed validation of this capability using the KITTI dataset~\cite{kitti}, as shown in Figure \ref{fig:app_LiDar}.

The ``Raw Inpainted'' column shows the input provided to our network. This is not the raw, sparse LiDAR point cloud itself, but rather a dense depth map created by applying a standard depth completion (inpainting) algorithm. While dense, this inpainted data suffers from two major problems: (1) significant artifacts from the completion process, such as horizontal streaking and noise, and (2) poor geometric definition, especially for distant or moving objects where the original point cloud was thinnest.

VDPP demonstrates its effectiveness as a powerful spatio-temporal refinement module for this noisy sensor data. It actively filters the inpainting noise while preserving the underlying geometric signal. In areas where the inpainted map shows blurred or poorly defined structures (like vehicle boundaries or distant scenery), VDPP refines the geometry, sharpens edges, and improves the depth separation between objects.

This validation on real sensor data confirms that VDPP is a truly scalable and practical solution. It can be seamlessly integrated into existing robotics stacks, not only to stabilize monocular depth estimation but also to serve as a high-performance refinement layer for noisy data from LiDAR or ToF sensors, significantly improving the quality and reliability of the perception system.

\section{Memory Efficiency Validation on Edge Devices}
\label{app:jetson}

To thoroughly evaluate VDPP's deployability on resource-constrained hardware, we conducted extensive memory profiling experiments on the NVIDIA Jetson Orin Nano (8GB), a widely-used edge computing platform for robotics and autonomous systems.

\subsection{Experimental Setup}
All experiments were conducted under identical conditions: Ubuntu 20.04, CUDA 11.4, with system idle memory baseline at 2.4 GB. We monitored memory usage using \texttt{jtop} and system resource monitors in real-time during inference from the Sintel dataset\cite{sintel} at native resolution.

\subsection{Successful Deployment Analysis}
In sharp contrast, Figure~\ref{fig:jetson_success} (main paper) captures the state of the system during sustained, real-time VDPP operation. This success is directly attributable to our geometric downsampling, which processes a compact manifold rather than full-resolution attention maps. By doing so, VDPP avoids the critical failure point identified in the failure case: the allocation of contiguous memory blocks that exceed device capacity. Crucially, it is worth noting that our deployment on the edge device does not rely on any additional model compression or optimization techniques, such as knowledge distillation or weight quantization. VDPP achieves practical inference speeds out-of-the-box. This demonstrates the inherent efficiency and 'plug-and-play' scalability of our framework, making it highly suitable for real-world applications where complex post-training optimizations are often costly or degrade baseline accuracy.

\subsection{Failure Case}
\begin{figure}[htbp!]
\centering
\includegraphics[width=\linewidth]{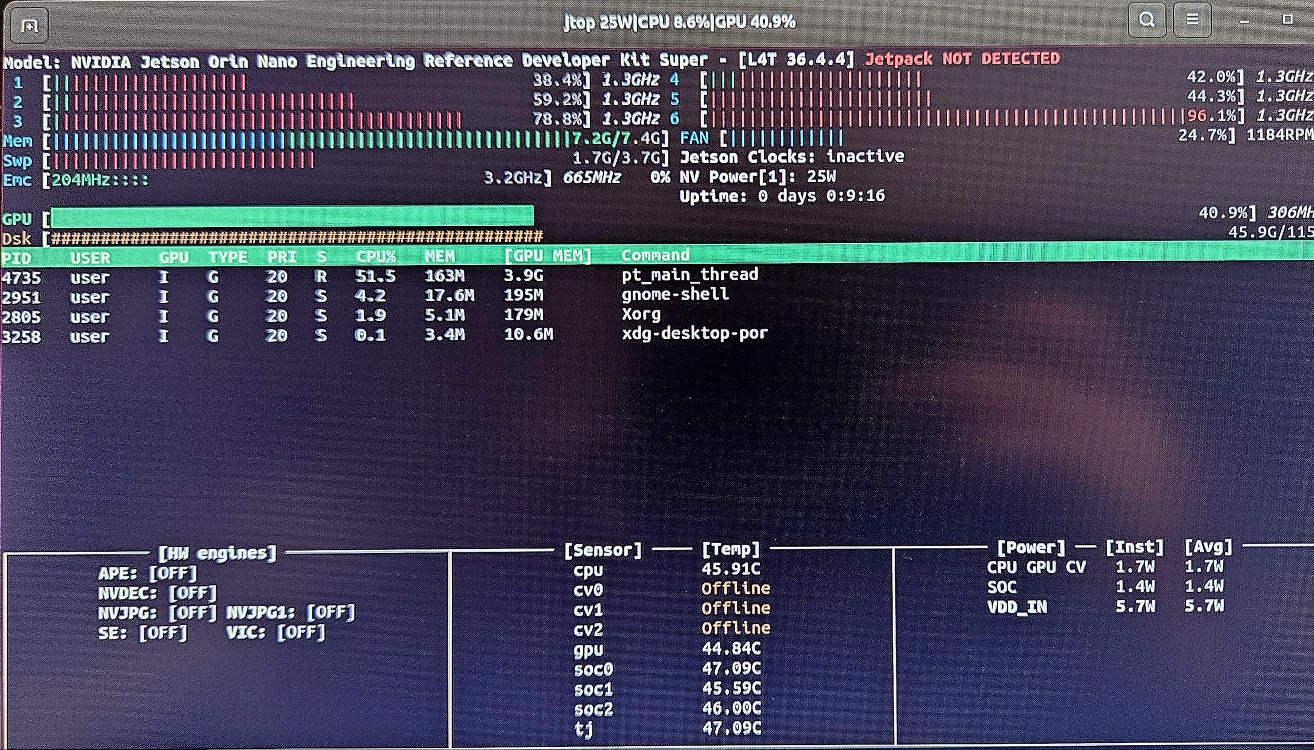}
\caption{Failed deployment attempts of VDA-L, VDA-S, and NVDS on NVIDIA Jetson Orin Nano, showing Out-Of-Memory errors and system crashes.}
\label{fig:jetson_fail}
\end{figure}

Figure~\ref{fig:jetson_fail} screenshots systematic failures of competing methods and shuts down the device. It Crashed at initialization when allocating attention matrices, estimated memory requirement (VDA-L, VDA-S). And VDPP with out downsampling also crashed at initialization, confirming downsampling is critical. Profiling reveals that full-resolution attention mechanisms allocate contiguous memory blocks that exceed device capacity, making them fundamentally incompatible with edge deployment regardless of optimization efforts.

These results demonstrate a categorical difference: VDPP enables a class of applications (edge-based video depth) that existing methods cannot support. The 2 GB memory headroom also allows co-deployment with other perception modules, critical for practical autonomous systems.

\section{Analysis of the TGSE Metric}
\label{app:tgse}
\begin{figure*}[ht!]
\centering
\includegraphics[width=1.0\linewidth]{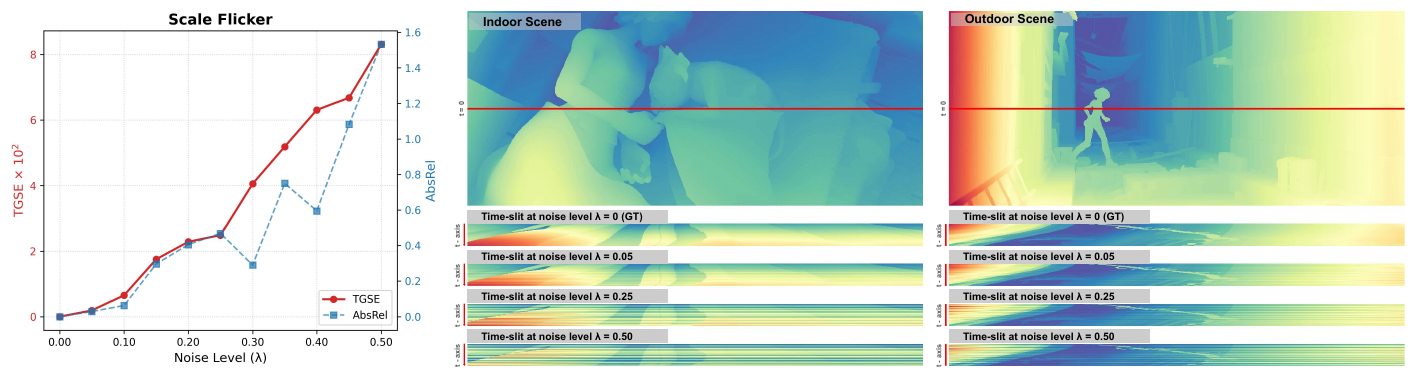}
\caption{\textbf{Sensitivity and Qualitative Analysis of Scale Flicker.} 
We analyzed metric behavior and visual quality changes while increasing the intensity of Scale Perturbation ($\lambda$) from 0\% to 50\%.
\textbf{Quantitative Sensitivity (Left):} The graph shows metric changes according to perturbation levels. Our \textbf{TGSE (\red{red line})} shows a strict \textbf{monotonic sensitivity}, accurately reflecting artifact severity. Conversely, \textbf{AbsRel (\blue{blue line})} fluctuates irregularly and exhibits unreliable behavior, specifically showing \textbf{a sharp drop at perturbation level 0.3 despite the intensified flicker.}
\textbf{Qualitative Temporal Profiles (Right):} Visualization of temporal profiles (slit-scans) for representative \textbf{Indoor (left)} and \textbf{Outdoor (right)} scenes from the Sintel dataset. The top of each block is the Reference Depth Map, with a red horizontal line indicating the slice position. The bottom stack shows temporal slices for Ground truth, $\lambda$ = 0.05, $\lambda$ = 0.25, and $\lambda$ = 0.5 (perturbation levels) in order. As perturbation increases, temporal discontinuities in the form of \textbf{horizontal banding} become evident. TGSE strongly correlates with this visual collapse, while AbsRel fails to capture it consistently.}
\label{fig:tgse}
\end{figure*}

In this paper, we proposed a novel metric, \textbf{Temporal Gradient Squared Error (TGSE)}, to evaluate the temporal consistency of video depth estimation. We conducted a sensitivity analysis in a controlled environment to demonstrate that existing spatial accuracy metrics (e.g., AbsRel, $\delta_1$) fail to properly capture temporal inconsistencies. This result highlights the necessity of a dedicated temporal metric and validates the reliability of TGSE.

\subsection{Experimental Setup}
We injected artificial temporal artifacts into the Ground Truth depth maps of all scenes in the Sintel dataset~\cite{sintel}, which possesses perfect temporal consistency, to observe how the metrics respond. To simulate `Flicker', the most common and detrimental artifact in video depth estimation, we applied \textbf{Frame-wise Multiplicative Scale Perturbation}.

For each frame $t$, we generated a perturbed depth $\tilde{D}^{(t)}$ by multiplying the Ground Truth depth $D_{GT}^{(t)}$ by a random scalar $s^{(t)}$:

\begin{equation}
    \tilde{D}^{(t)} = D_{GT}^{(t)} \cdot s^{(t)}, \quad \text{where } s^{(t)} \sim \mathcal{U}(1-\lambda, 1+\lambda)
\end{equation}

where $\mathcal{U}$ denotes a uniform distribution, and $\lambda$ represents the perturbation level. We incrementally increased $\lambda$ from $0.0$ to $0.5$ and measured the changes in TGSE and AbsRel.

\subsection{Quantitative Analysis}
An ideal temporal consistency metric should exhibit a monotonic increase in error value as the perturbation intensity ($\lambda$) increases. Figure~\ref{fig:tgse} (Left) presents the results of this experiment.

\noindent \textbf{Robustness of TGSE:} Our TGSE (\red{solid red line}) maintains a strict monotonic increasing trend as the perturbation level rises. This demonstrates that TGSE accurately quantifies the intensity of flicker artifacts. The temporal gradient amplifies the variance of independent perturbation noise between frames by a factor of two, facilitating detection. Furthermore, the L2 Norm imposes a high penalty on large outliers, preventing score degradation due to stochastic effect.
    
\noindent \textbf{Blindness of Spatial Metrics:} In contrast, the standard spatial metric, AbsRel (\blue{dashed blue line}), shows unpredictable fluctuations. Most critically, despite the perturbation level increasing from 0.25 to 0.30, the AbsRel value paradoxically drops sharply. This indicates that spatial metrics are vulnerable to stochastic alignment of per-frame pixels independently and may fail to correctly assess the degradation of temporal quality.

\subsection{Qualitative Analysis}
Figure~\ref{fig:tgse} (Right) visualizes the temporal profile (slit-scan) of the experiment. As the perturbation level increases from 0 to 0.5, discontinuities in the form of \textbf{horizontal banding} become more pronounced in the temporal slice.

TGSE scores increase in strong correlation with this temporal degradation, whereas AbsRel fails to consistently reflect it. Therefore, a dedicated temporal metric like TGSE is essential alongside spatial metrics for the fair evaluation of video depth estimation models.

\clearpage


\end{document}